# On the Design of Agent-Based Systems using UML and Extensions

Mihaela Dinsoreanu, Ioan Salomie, Kalman Pusztai
*Technical University of Cluj-Napoca*
*Computer Science Department*
*Baritiu Str. 26-28*
*RO-3400, Cluj-Napoca, Romania*
*{Mihaela.Dinsoreanu, Ioan.Salomie, Kalman.Pusztai}@cs.utcluj.ro*

**Abstract**. *The Unified Software Development Process (USDP) and UML have been now generally accepted as the standard methodology and modeling language for developing Object-Oriented Systems. Although Agent-based Systems introduces new issues, we consider that USDP and UML can be used in an extended manner for modeling Agent-based Systems.*

*The paper presents a methodology for designing agent-based systems and the specific models expressed in an UML-based notation corresponding to each phase of the software development process. UML was extended using the provided mechanism: stereotypes. Therefore, this approach can be managed with any CASE tool supporting UML. A Case Study, the development of a specific agent-based Student Evaluation System (SAS), is presented.*

## 1. Introduction

The Unified Software Development Process (USDP) and UML have been now generally accepted as the standard methodology and its associated modeling language for developing Object-Oriented Systems. Although Agent-based Systems introduces new issues like autonomy, reasoning, mobility etc., we consider that USDP and UML can be used in an extended manner for designing agent-based systems.

There are several reasons for choosing USDP and UML for specifying requirements and designing agent-based systems: first, in the requirements specification and early analysis phases the focus stays on the functionalities of the system and not on the technologies used for implementing the system.

Therefore, the use of agents or another component type approach is transparent at this moment to the analyst and does not influence the requirements specification. Second, USDP and UML provide a widely known and used methodology for specifying requirements and modeling use-cases. In the design phase where agent-specific issues come up, the mechanism for extension provided by UML can be exploited: stereotypes can be used for capturing agent-specific concepts. Last but not least, UML is supported by a number of computer-aided software engineering platforms.

This paper presents a methodology for designing agent-based systems and the specific models expressed in an UML-based notation corresponding to each phase of the software development process. A specific case study is presented: the development of a Student Assessment System (SAS), integrated in a Virtual University. We present the software engineering methodology used and also the corresponding model created using UML and extensions. SAS is a closed, dynamic, distributed agent-based system containing benevolent agents, both stationary and mobile, that cooperate in order to provide an efficient and reliable assessment service in a Web-based Distance Education Environment.

Section 2 outlines the methodology in terms of the necessary phases for the development of an agent-based system. For each phase the corresponding models are presented. Section 3 provides a more detailed presentation of the models specified in the previous section and also their representation using UML and extensions. Section 4 presents a case study related to a specific agent-based system, the SAS. Some of the diagrams elaborated for modeling the SAS are discussed. Section 5 outlines some conclusions and future development intentions.

## 2. Methodology

Since agent-based systems became more and more used not only in the academia but also in practical industrial environments, a methodology for developing such systems became more important. Different methodologies were developed and proposed, all of them being inspired by older software engineering methodologies. There are two main starting points:

- The Object oriented approach, having UML as support language.
- Knowledge Engineering approach.

The approach adopted in our work under development uses as starting point the Object-Oriented methodology and UML.

Our approach does not consider the multi-agent system as a set of independent agents that act individually in the same environment in order to fulfill some goals. We consider multi-agent systems a set of interrelated agents that communicate to each other trying to fulfill not only their individual goals but also common goals. The agents should also respect some social laws – forming therefore a society or an organization.

As stated above, the methodology is based on previous work [3], [8], [9] being a top-down approach that starts analyzing the functional requirements of the system considered an organization. The methodology contains a requirements specification phase, an analysis phase and also a design phase. The implementation and deployment phases are not discussed here.

The main steps of the methodology are presented below:

- Specification of the **functionality** of the system as an organization. The functionality of the system is expressed in terms of provided services to the user. The provided services can be mapped on the **social tasks** of the agent organization members.
- Designing the **organizational model** in terms of the **role model** and **social rules**. This means the identification of positions/functions that must be fulfilled by agents besides their individual tasks and also the social laws that should be obeyed by all the agents members of the organization.
- Identifying the **interaction patterns** that represent communication protocols between different roles, therefore defining the **interaction model.**
- Designing the **environment model** in terms of existing resources and access protocols to them.

In the design phase the concepts defined above are refined on a lower abstractization level.

The main steps are presented below:

- Designing the **agent model** that should support the role model and interaction model defined above.
- Designing a **coordination model** that contains the coordinated elements, a coordination media and also a set of coordination laws. This model is derived from the interaction model and the set of social rules identified above.
- Designing the **services** provided by the agents. The **services model** should consider both the individual agent tasks and also the social tasks of the entire organization.

## 3. Modeling using UML

The **functionality** specification phase can be approached in the classical USDP way, by identifying the external actors that interact with the system and the requested functionalities modeled as packages in UML.

### 3.1. Analysis models

The **organizational model** consists of a role model and a model of the social rules.

The **role model** represents positions/functions that will be fulfilled by specific agents. A role is defined in terms of the individual tasks it has to perform, of the possible interactions, of the resources it has access to and of the social tasks it is involved in. Since there is no UML representation for roles/agents, we used stereotypes representing roles as "business worker" in UML. The tasks a role should perform are represented in UML as use-cases attached to a role. The Use-cases that involve agent roles are represented by "business use-cases". The interaction patterns are represented as sequence/collaboration diagrams. The social rules are modeled as constraints that will be applied on the interactions between roles.

The **environment** model will represent the available resources provided to the agent roles. These are represented in UML as regular entity classes. Access protocols to the resources are also represented using sequence/collaboration diagrams.

### 3.2. Design Models

Concerning the design phase, we will focus on modeling the agent behavior. According to [4] agent behavior can be modeled as a number of Concurrent Tasks. These tasks specify a single thread of control that defines one task an agent can perform and contains both inter-agent and intra-agent interactions.

Each of these tasks executes in parallel to define the behavior of the agent. The behavior is defined in terms of planning and plan implementation and not in terms of the individual plans themselves.

The approach adopted for modeling concurrent tasks is by using statechart diagrams. The statechart diagrams contain two main elements: states and transitions. According to [3] states encompass the processing that goes on internal to the agent while transitions allow communication between agents. A transition consists of a source state, destination state, trigger, guard condition and transmissions. The general syntax for a transition is:

*trigger [guard] transmission*

A trigger can be either a message received from another agent or an internal event that occurred during another task. Transmissions are either messages sent to external agents or events sent to another internal task. Guards are boolean conditions that should be true before the transition takes place.

For representing messages that are sent between agents two special events are used: *send(message, agent)* and *receive(message, agent)*. Messages to a group of agents can be sent via multicasting.

States may contain activities that represent internal reasoning, reading perceptions from sensors, performing actions via effectors etc. Multiple activities may be included in a single state and are performed sequentially. Once in a state, the task remains in that state until all the activities are carried out and a transition out of the state becomes enabled.

Activities are defined as functions having a set of input parameters and returning up to one result.

Dealing with mobility and time was also foreseen. Mobility is managed using the *move* activity. The syntax for this activity is

*Boolean = move(location)*

where *location* denotes the destination location of the agent. For reasoning about time, the model provides a built in timer activity that can be set using the *settimer(time)* activity and can be tested using the *timeout(t)* activity. Actually, the *timeout* activity is not generally used in a state, but as a guard condition on transitions.

### 3.3. Formal Verification

Concurrent tasks modeled as statechart diagrams can be formally verified in order to detect communication centric errors like: deadlocks, livelocks, assertion violations and others. A methodology for a formal verification [5], involves translation from the graphical, UML-based statechart diagrams into a formal language called Promela. The formal Promela model can be verified using another tool called Spin. Our work under development aims to develop a verification tool to automatically translate UML-based statechart diagrams into Promela models and to apply Spin on the formal model.

### 4. Case Study

Considering the approach shortly presented above, our work under development aims to model an agent-based system, able to provide assessment services for students enrolled in a Virtual University (VU). VU is a Web-based Distance Education Environment, developed at the Computer Science Department of the Technical University of Cluj-Napoca, in order to provide web-based support for Distance Education programs. We aim to integrate in VU an agent-based module providing assessment services to the student enrolled.

In order to define the functionalities of the system we identified two main functional approaches:

- A Pull (Self-Assessment) Scenario initiated by the Student, who learned a certain section of a specific matter and wants to evaluate his/her knowledge. In this case the Test type is configured by the Student and no record of the assessment is registered in the VU.
- A Push (Exam) Scenario initiated by the Teacher, who enforces a certain Test type for evaluating the students' knowledge level. In this case the configuration is done by the Teacher and the result of the evaluation is recorded in VU.

A very general UML description of the two main approaches is depicted in Fig 1.

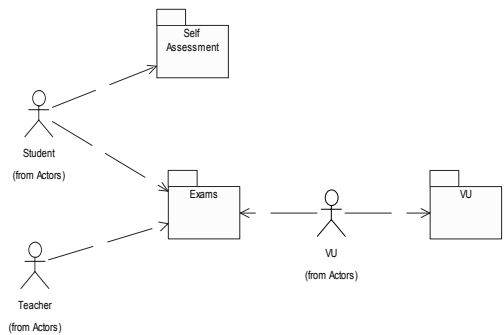

**Figure 1. Main functional modules**

We identified three external actors of the system: the Student, the Teacher and the VU.

We will consider the Self-Assessment module, the Exam module being treated the same way. The Self-assessment module can be further detailed considering the sub-modules depicted in Fig 2. This way we can identify some of the social tasks of our system like Assistance, Exam Generation, Taking Exam etc.

For accomplishing the Assistance task we considered the need of a Personal Assistant Agent (PAA) providing an interface for the student to interact with the system. The PAA would be a stationary agent residing on the student's machine. He interacts on the other side with the SAS, communicating the student's requests and providing access to the local resources for the assessment.

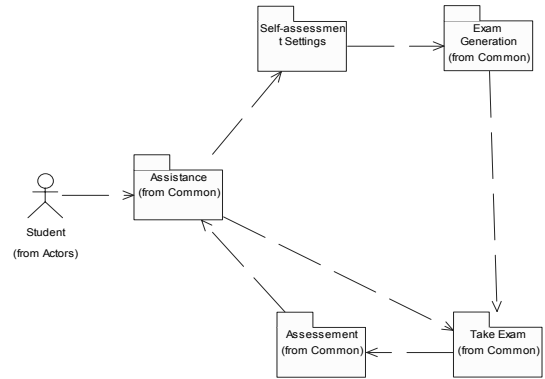

**Figure 2. Sub-modules for Self-assessment**

In the Logical View the structure of the Assistance module is represented like in Fig 3.

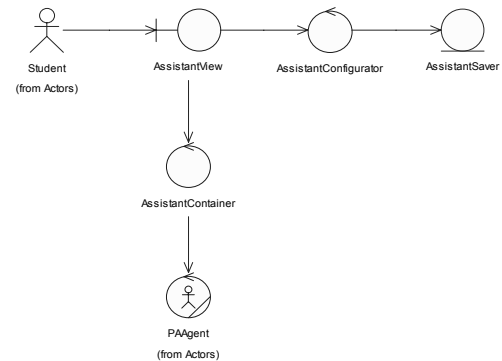

**Figure 3. Logical View of Assistance**

We considered on the SAS side the need of a Server Agent (SA) role. The SA should be responsible with managing PAA's requests, initiating the creation of a specific Evaluation Engine corresponding to the configuration received and also initiating the creation of an Evaluation Agent (EA) responsible with the actual evaluation. SAs are also static agents residing on the SAS's site.

The evaluation has two main phases:
- an offline phase where specific domain knowledge is acquired creating the domain knowledge base. In this phase also the expert answers of the test are analyzed and structured.
- an online phase where the students' answers are analyzed and matched against the expert answers structures.

The off line phase takes place before any assessment is performed.

The need of other components of the SAS is obvious: an Evaluation Engine Factory that should create specific Evaluation Engines for specific assessment configurations. The Evaluation Engine is attached to an EA and provides its ability to analyze the student's answer and to match it against the expert answer, therefore being able to evaluate it.

Another important component is an Agent Factory that actually creates EA's. The EA is a mobile agent, loaded with assessment knowledge (the Evaluation Engine), with a set of questions and expert answers. The EA travels to the student's site and co-operates with the PAA in order to get the assessment done. The EA has an adaptive behavior depending on the student's answers.

As shown in Fig. 4 one of the responsibilities of the SA is to manage Assessment requests received from the PAA and to request the creation of a corresponding Evaluation Engine and of an EA. These two tasks are represented as attached use-cases to the "business worker" SA.

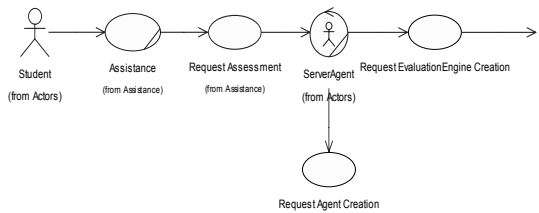

**Figure 4. Exam Generation module**

According to the adopted methodology another modeled aspect would be the interaction between the elements of the SAS. We used UML Sequence Diagrams for modeling the interactions.

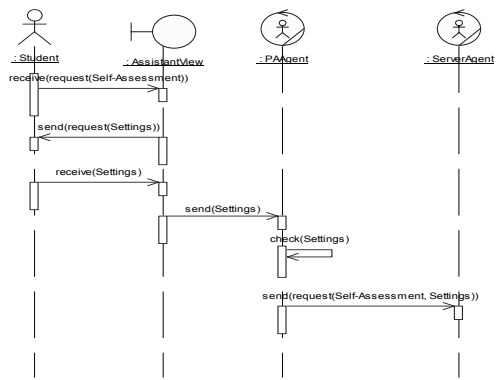

**Figure 5. Self-Assessment Request Interaction**

As shown in Fig 5. the interactions involved in the Self-Assessment Request can be clearly represented using Sequence Diagrams.

We will not focus in this paper on the structure of the Evaluation Engine. The Evaluation Engine will be able to manage different test types like: multiple choice tests, short answers using natural language etc. using a natural language processor based on latent semantic analysis approach.

The AgentFactory was modeled using a creational design pattern: Abstract Factory.

Fig 6. represents an example of such a statechart diagram that models the behavior of the Evaluation Agent in the Take Exam module.

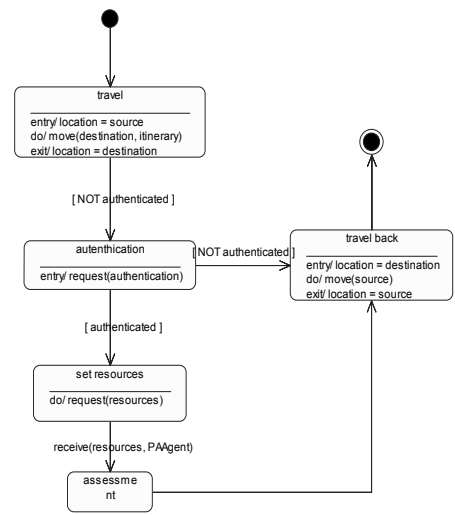

**Figure 6. Concurrent Task**

The statechart diagrams are to be formally verified by generating a formal model using Promela as the formal representation language and Spin as the verification tool [5].

The VU is developed, tested and used at the Computer Science Department of TUCN. The technologies used are Java Technologies (JSP, JavaBeans) and MSSQL database server. SAS is currently under development.

## 5. Conclusions

Based on the models presented, we believe that the methodology developed and used can be considered a foundation for designing and developing multi-agent systems.

It takes advantage of a goal-driven approach, considers agent-specific issues like roles, tasks and interactions in the analysis phase and can be supported by a well-known modeling language as UML, therefore several CASE tools like Rational Rose being appropriate to be used.

Of course, the methodology and the models need further work and development. One of the issues not addressed would be modeling cooperation and coordination. The current approach uses interaction diagrams for these models, too.